\documentclass[10pt,journal,compsoc]{IEEEtran}

\usepackage{graphicx}
\usepackage{booktabs}       % professional-quality tables
\usepackage{amsfonts}       % blackboard math symbols
\usepackage{nicefrac}       % compact symbols for 1/2, etc.
\usepackage{microtype}      % microtypography
\usepackage{amsmath}
\usepackage{amssymb}
\usepackage{gensymb}
\usepackage{amsthm}
\usepackage{adjustbox}
\usepackage{subfigure}
\usepackage{xcolor}
\usepackage{algorithm}
\usepackage{algpseudocode}
\newtheorem{theorem}{Theorem}

\ifCLASSINFOpdf
\else
\fi
\begin{document}
\title{Privacy-Preserving Constrained Domain Generalization via Gradient Alignment}

\author{Chris Xing Tian, Haoliang Li, Yufei Wang and Shiqi Wang% <-this % stops a space
%\thanks{M. Shell was with the Department
% of Electrical and Computer Engineering, Georgia Institute of Technology, Atlanta,
% GA, 30332 USA e-mail: (see http://www.michaelshell.org/contact.html).}% <-this % stops a space
% \thanks{J. Doe and J. Doe are with Anonymous University.}% <-this % stops a space
% \thanks{Manuscript received April 19, 2005; revised August 26, 2015.}
}


\markboth{Journal of \LaTeX\ Class Files,~Vol.~14, No.~8, August~2015}%
{Shell \MakeLowercase{\textit{et al.}}: Bare Demo of IEEEtran.cls for IEEE Journals}



\IEEEtitleabstractindextext{%


% as well as two challenging medical imaging classification tasks

%For example, in the field of medical imaging classification, when training the DNN from one domain (e.g., with data only from one hospital), the generalization capability to another domain (e.g., data from another hospital) could be largely lacking.

\begin{abstract}
Deep neural networks (DNN) have demonstrated unprecedented success for various applications. However, due to the issue of limited dataset availability and the strict legal and ethical requirements for data privacy protection, the broad applications of DNN (e.g., medical imaging classification) with large-scale training data have been largely hindered, greatly constraining the model generalization capability.  In this paper, we aim to tackle this problem by developing the privacy-preserving constrained domain generalization method, aiming to improve the generalization capability under the privacy-preserving condition. 
In particular, we propose to improve the information aggregation process on the centralized server side with a novel gradient alignment loss, expecting that the trained model can be better generalized to the ``unseen" but related data. The rationale and effectiveness of our proposed method can be explained by connecting our proposed method with the Maximum Mean Discrepancy (MMD) which has been widely adopted as the distribution distance measure.  Experimental results on three domain generalization benchmark datasets  indicate that our method can achieve better cross-domain generalization capability compared to the state-of-the-art federated learning methods.
\end{abstract}

}

\maketitle

% \fi
\IEEEpeerreviewmaketitle

\section{Introduction}

%With the development of artificial intelligence (AI) technology, 
Deep neural networks (DNNs) have achieved remarkable success in various applications, such as computer vision, natural language processing, and acoustic verification. For example, in the field of medical imaging classification (e.g., tumour detection and classification, X-ray image analysis), DNN model can even achieve higher diagnosis accuracy compared with human doctors. 

The tremendous achievements of DNNs are driven by the availability of the large-scale training data. To guarantee reliable decision support based on artificial intelligence (AI) oriented applications, the large amount of data are indispensable for training purposes \cite{najafabadi2015deep,kaissis2020secure}. However, reasonably large-scale dataset for some realistic environments (e.g., clinical environment \cite{guan2021domain, yang2022minimally}) are infeasible to collect due to two reasons. First, though annotated data can be collected from multiple sources, aggregating the data for training may not be feasible due to the privacy regulations. For example, European General Data Protection Regulation (GDPR) has imposed strict rules regarding the storage and exchange of the health data. Second, even though data access permission can be obtained, collecting large-scale and representative data can still be difficult due to the variation of capturing protocols, device vendors and environments. Thus, the trained DNNs are typically prone to be lack of generalization capability when annotated large-scale data are not available during training stage, especially for the out-of-distribution data which are ``unseen" during the training stage. 

Tremendous efforts have been devoted to tackling the challenges of privacy and generalization for the DNN. Regarding the issue of privacy, federated learning (FL) \cite{yang2019federated,mcmahan2017communication} was proposed to train DNN based on datasets distributed across multiple domains while preventing data leakage. However, while existing techniques (e.g., \cite{li2019privacy,li2021fedbn}) can tackle the setting where data from multiple domains are heterogeneous (i.e., the distribution of data from different sources are different), the trained DNN may not be able to generalize well to the out-of-distribution data.  Regarding the issue of generalization, numerous methods have been developed to improve the generalization capability of DNN based on domain generalization \cite{dou2019domain,li2020domain}. However, the existing techniques require to aggregate the data to conduct domain-shift simulation, which disobeys the privacy regulation rules. 

To jointly overcome the aforementioned difficulties, we propose a novel \textit{task-agnostic} domain generalization method based on gradient aggregation, aiming to improve the model generalization capability under the privacy-preserving constraint (i.e., domain generalization under the constraint of privacy-preserving without data sharing from multiple domains). By treating the gradient as a kernel mean embedding from the original data space to the neural tangent kernel space, we conduct distribution alignment through Maximum Mean Discrepancy (MMD) \cite{gretton2012kernel} across multiple domains based on the gradient. As such, the new aggregated gradient is equipped with information from multiple domains and is expected to be better generalized to the ``unseen" testing data. We conduct extensive experiments on three domain generalization benchmarks with the issue of privacy  to evaluate our proposed method. The results show that our method can achieve better generalization capability compared with state-of-the-art FL and domain generalization (DG) techniques under the privacy-preserving setting.  
% The contributions of our papers are summarized as follow.
% \begin{itemize}
%     \item We propose to tackle the problem of domain generalization under a privacy-preserving constrained setting. Particularly, by treating the gradient information as representative information for each domain (i.e., local server), a novel information aggregation module on the central server side is proposed where we expect the trained DNN can be better generalized to out-of-distribution samples. 
%     \item We theoretically prove the training convergence of our proposed information aggregation module, which  shows the feasibility of our proposed method for practical deployment. 
% %    \item The experimental results based on various tasks show that our proposed method can achieve
% better performance compared with the state-of-the-art methods based on  privacy-preserving cross-domain recognition problems.

% \end{itemize}

\section{Related Works}

\subsection{Domain Generalization}
% To tackle the problem where training and test data are collected under different environmental conditions, domain generalization is proposed to use data collected from various environmental conditions to train a DNN model to handle testing data collected from an unseen but related condition. Existing techniques can be categorized into two streams: (1) extracting shareable information across data through feature representation learning (e.g., \cite{muandet2013domain}) and meta-learning (e.g., \cite{li2018learning}); and (2) conducting data augmentation (e.g., domain randomization \cite{tobin2017domain}, adversarial training \cite{volpi2018generalizing,zhou2020deep}) to enhance the scale of training data. Whereas domain generalization has been proven to be effective in various AI tasks, its use in privacy-preserving conditions has yet to be explored. Therefore, we aim to tackle the problem of privacy-preserving domain generalization in our paper.  

{In the context of mitigating the challenge posed by disparate environmental conditions between training and test data, the concept of domain generalization (DG) has emerged as a promising approach. DG involves leveraging data collected from diverse environmental conditions (source domains) to train a deep neural network (DNN) model that can effectively handle testing data obtained from an unseen yet related condition (unknown target domain). It is important to note that DG shares a close relationship with domain adaptation (DA) \cite{wang2018deep}, where domain shifts are also addressed. However, unlike DA, which assumes access to some (labeled or unlabeled) data samples from the target domain, DG assumes the unavailability of such samples during training. Consequently, DG methods must seek solutions to effectively exploit the information present in multiple source domains accessible during training. The hope is that by distilling shared knowledge from source domains, we can obtain more robust features that can be potentially useful in unseen target domains. Existing techniques in the DG field can be broadly categorized into three streams.}
{The first stream is based on the idea of training a dedicated classifier for each source domain and then combining them to give fused predictions by evaluating the similarity between different source domains and test samples (e.g.,  \cite{xu2014exploiting, mancini2018best}).}
The second stream focuses on extracting shareable information across data through feature representation learning (e.g., \cite{muandet2013domain}) and meta-learning (e.g., \cite{li2018learning}). These approaches aim to discover common patterns or representations that are transferable across different domains, enabling the model to generalize well to unseen conditions.
The last stream involves employing data augmentation techniques, such as domain randomization \cite{tobin2017domain} and adversarial training \cite{volpi2018generalizing,zhou2020deep}, to augment the scale of the training data. By introducing variations or perturbations to the data during training, these methods enhance the model's ability to handle diverse environmental conditions.

\subsection{Federated Learning}
Recent years have seen a rapidly growing number of intelligent devices with AI computing capability, such as smartphones, wearable devices, autonomous vehicles, intelligent CCTV cameras, IoT devices, etc. Those devices, forming a large distributed network, can generate a large amount of heterogeneous data every day. How to fully utilize the local AI computing capability of each device while reducing the data transmission cost or preserving data privacy becomes a new challenge. Traditional AI data processing models, which usually require homogeneous data transmission from some parties to a central party for model training and final build, can hardly be adapted in such scenarios. This gap leads to a growing interest in the Federated Learning (FL) framework. The Federated Learning is firstly proposed in \cite{konevcny2016federated,mcmahan2017communication} to support training AI models over distributed remote devices or isolated data centers while keeping data localized. In a general Federate Learning setting, there may be tens to potentially millions of distributed clients (remote devices/soiled data islands, etc.), and each client trains the AI model locally using its private dataset. In each FL training round, the clients will share their model information, usually the learned model weights or gradients, instead of the training data, with a central aggregator. The aggregator would aggregate the information from those clients (e.g., through model parameters averaging) to obtain global model parameters, which will be sent back to clients for the next round of training.

    One limitation of the aforementioned mechanism is that it does not fully address the underlying challenges associated with system and data heterogeneity, where system heterogeneity refers to the situation where each local server has different computational power, communication bandwidth, etc., which further leads to local-update variation, and data heterogeneity refers to the situation that the data distributions from different local servers are different. Moreover, it is highly likely that the testing data are drawn from the distribution which is different from the data distributions of local servers. Regarding the first issue, in \cite{li2019privacy}, a proximal term on the objective function is introduced for each local server, such that the impact of local-update variation can be mitigated. Regarding the second issue, in \cite{gao2019privacy,liu2020secure}, a federated transfer learning scheme was proposed, where shareable information across servers can be learned with domain alignment regularization. However, it required that the co-occurrence samples are available between the labeled source domain (i.e., domain for training purpose) and the unlabeled target domain (i.e., domain for testing purpose), which is not desired as we do not have target domain data in hand for real-time applications. As such, its generalization capability to out-of-distribution samples is prohibited. Besides the aforementioned techniques, there are also some works focusing on the federated learning for multi-task setting or non i.i.d setting \cite{smith2017federated,yu2020learning,luo2021no,yue2022neural}, where the data are heterogeneous. However, they are not designed based on the cross-domain scenario for testing.  

    In recent years, there exist some works which focus on tackling the problem of domain generalization under the constraint of privacy-preserving (i.e., non-shared data from multiple domains) \cite{liu2021feddg,wu2021collaborative}. In \cite{liu2021feddg}, the authors proposed to tackle the problem domain generalization under privacy setting for medical imaging, with the input based on frequency space (with 2D Fourier transformation) and a boundary-oriented meta-optimization strategy, which is  task-specific (i.e., tailored for medical image segmentation task). While the method proposed in \cite{wu2021collaborative} is task-agnostic, it is built upon FedAvg \cite{mcmahan2017communication} but requires both trainable classifier models and frozen models during the training stage, which may not be able to scale to different architectures. Our framework only requires conventional local training on clients, which aligns with the standard of federated learning and can be applied to different models and tasks. Moreover, unlike \cite{liu2021feddg} and \cite{wu2021collaborative} which focus on local training, our proposed method focuses on aggregation on the central server side.

\section{Proposed Method}

We propose to study the problem of task-agnostic privacy-preserving constrained domain generalization (PPDG). The architecture is built upon the FL system configuration \cite{yang2019federated,mcmahan2017communication}, which is designed to handle data from multiple local servers (i.e., client servers) and then aggregate the information from local servers to a centralized server. Based on the FL settings, the centralized server maintains a global DNN model to coordinate the global learning objective across the framework. Specifically, the objective is to minimize
\begin{equation}
\min_{w} f(w) =\sum_{k=1}^{K} p_k F_k(w), 
\end{equation}
where $F_k(w)$ denotes the objective of deep learning model on the $k$-th local server, $K$ is the number of local servers, $p_k>0$, and $\sum_k p_k=1$. In practice, one can set $p_k = n_k/n$, where $n_k$ and $n$ denote the number of training data in the $k$th server and the total number of training data, respectively. During training, at the federated round $t$, the DNN in the local server is updated by receiving the DNN parameters/gradients\footnote{We consider gradient in our manuscript.} from the centralized server, and the local servers further conduct DNN model training and send the encrypted gradient to the centralized server for gradient aggregation.

\subsection{Distribution Alignment in Neural Tangent Kernel Space}

Directly conducting gradient aggregation through averaging process \cite{mcmahan2017communication} may not benefit the generalization capability of DNN model. One reason may be attributed to the gradient conflict (i.e., $\langle \frac{\partial F_i(w)}{\partial w},  \frac{\partial F_j(w)}{\partial w} \rangle <0$ for local server $i$ and $j$) \cite{riemer2018learning} which further leads to negative transfer across different servers. To this end, we propose to improve the generalization capability of DNN with privacy-preserving constraints by proposing a novel gradient aggregation technique. 

Before introducing our proposed method, we first revisit the problem of improving the generalization capability of machine learning model to the out-of-distribution samples, which has received more and more attention recently \cite{muandet2013domain,li2018domain}. In \cite{li2018domain}, the authors theoretically proved that the generalization capability can be improved by domain alignment through domain variance reducing, where the domain variance can be defined by summing the MMD distance between domain pairs, which can be given as 
\begin{equation}
\sum_{i,j} \|\mu_{P_i} - \mu_{P_j}\|_\mathcal{H}^2, 
\end{equation}
where $\mu_{P_i}$ and $\mu_{P_j}$ denote the kernel embedding of distribution of domain $i$ and $j$, respectively. One can minimize the domain variance by representing the kernel mean with empirical averaging as 
$\mu_{P} = \frac{1}{n} \sum_{i=1}^{n} \phi(x_i)$,
where $\phi$ denotes a feature mapping function \cite{gretton2012kernel}.  

Our motivation originates from the recent advance of DNN analysis based on neural tangent kernel \cite{jacot2018neural}. By conducting first-order Taylor expansion of network objective $f(w)$, we can reformulate the objective as 

\begin{equation}
    f(w)  \approx f(w_0) + \bigtriangledown_w f(w_0)^\top (w-w_0).
\end{equation}

By focusing on the parameter $w$, the above approximation can be interpreted as a linear model with respect to $w$, and the feature map $\phi(\cdot)$ is the gradient at the initialization $w_0$ given as $\phi(x) = \bigtriangledown_w f(x;w_0)$ w.r.t. the data $x$. Based on the neural tangent kernel space, where the feature map of the original input data $x$ is defined as the corresponding gradient, we can interpret the average gradient as the kernel mean with empirical averaging, given by $\mu_{P} = \frac{1}{n} \sum_{i=1}^{n} \phi(x_i)$, where $\phi$ represents the feature mapping function based on the neural tangent kernel, which is the gradient. Thus, we can define the domain variance among multiple local servers based on the kernel embedding in the neural tangent kernel space by extending the MMD distance in Eq. (2) as
\begin{equation}\label{eq:grad}
    \sum_{i,j} \|grad_i - grad_j\|^2,
\end{equation}
where $grad_i$ and $grad_j$ denote the gradient sent to the centralized server from local server $i$ and $j$, respectively.

\subsection{Gradient Aggregation}
\begin{algorithm}
\begin{algorithmic}[1]
\caption{Proposed Gradient Aggregation Algorithm}
\State \textbf{input:} Gradients sent from local servers: $G=\{grad_i\}$.

% \For{s = 0, 1, 2 ,...}\Comment Overall iterations
\State \textbf{Initialize:} $\hat{grad}_i = grad_i$, $G=\{\hat{grad}_i\}$
% \State Compute $\nabla{\mathcal F(w_s) = \frac{1}{K}\sum_{i=1}^{K}{\nabla{\mathcal F_i(w_s)}}}$
\For{$\hat{grad}_i \in G$ } 
% \State $w_i = w_i - \alpha \nabla \mathcal F(w_s)$
% \State Initialize: $ w_i = \tilde{w_s}$

\For{$\hat{grad}_j \in G \backslash \{\hat{grad}_i\}$} 
% \State $w_j = w_j - \alpha \nabla \mathcal F(w_s)$
% \State Initialize: $ w_j = \tilde{w_s}$
% \IF{$\left \langle \frac{\partial f_i(w_i)}{\partial w}, \frac{\partial f_j(w_j)}{\partial w} \right \rangle \textless 0 $}
\If{$ \langle \hat{grad}_i,\hat{grad}_j\rangle< 0 $}

\State $\hat{grad_i} = \hat{grad}_i - 2 \lambda (\hat{grad}_i - \hat{grad}_j)$ 

\Comment{Gradient alignment}
\EndIf
\EndFor 
\EndFor
\State $grad_{agg} = \frac{1}{K}\sum_{i=1}^{K} \hat{grad}_i$ \Comment{Aggregated gradient sent back to local servers}
\end{algorithmic}
\end{algorithm}

The gradient aggregation on the centralized server side  optimizes Eq.~\ref{eq:grad} instead of conducting gradient averaging to avoid possible gradient conflicting. To jointly achieve domain alignment among multiple local servers while preserving discriminative power of DNN learning, we propose to conduct gradient modification based on local server $i$ through gradient descend w.r.t. local server $j$ only if negative transfer between $i$ and $j$ occurs (i.e., $\langle \frac{\partial F_i(w)}{\partial w},  \frac{\partial F_j(w)}{\partial w} \rangle <0$), and the modified gradient of local server $i$ w.r.t. local server $j$ is given as
\begin{equation}\label{eq:gradient_update}
    \hat{grad_i} = grad_i - 2 \lambda (grad_i - grad_j),
\end{equation}
where $\lambda$ is the hyper-parameter for gradient descend. We repeat this process across the gradients collected from all local servers in a random order to obtain the respective gradient $\hat{grad_i}$ for local server $i$. We then conduct gradient averaging, which is for model update. Our proposed method is summarized in Algorithm 1.

It is worth noting that our proposed gradient aggregation does not conflict with the homomorphic encryption, which encrypts the data by preserving the structural transformation of original data \cite{yang2019federated,gentry2009fully} and is adopted for gradient communication in the FL setting.  For example, by considering homomorphic encryption, Eq.~\ref{eq:gradient_update} can be reformulated as
\begin{equation}
    \hat{E}(grad_i) = E(grad_i) \ominus E(2) \otimes E(\lambda) \otimes (E(grad_i) \ominus E(grad_j)), \nonumber
\end{equation}
where $\otimes$ and $\ominus$ denote the subtraction and multiplication operation in the encrypted space.  After conducting gradient aggregation on the centralized server side, the modified and encrypted gradient can be sent back to the local servers for decryption and model updating.

\textbf{Discussion. } Conceptually, our proposed method is close to Meta-Learning Domain Generalization (MLDG) \cite{li2018learning}. In \cite{li2018learning}, a first-order Meta-Learning approximation was proposed to reformulate the objective in \cite{finn2017model} with a classification loss and a gradient similarity loss. Besides, there also exists methods which perform gradient alignment for the problem of domain generalization. For example, ArgSum and ArgRand \cite{mansilla2021domain} aim to explore the consensus of gradient information (i.e., by masking the gradient to $0$ if there exists sign contradiction). While ArgSum and ArgRand are based on the setting where data from multiple domains could be shared, their method can be extended in our setting. However, these two methods only explore gradient sign information which may not be able to extract domain invariant knowledge across different clients. In \cite{shi2021gradient} and \cite{rame2022fishr}, the authors proposed to maximize the gradient similarity, and minimize gradient variance across domains, which can also be treated as performing gradient alignment for domain generalization. However, \cite{shi2021gradient} and \cite{rame2022fishr} focus on the centralized setting (i.e., data from multiple domains are shared), where the gradient alignment loss is jointly optimized with classification loss in a gradient-of-gradient manner (see Algorithm 2 in \cite{shi2021gradient}). Our proposed method is different from \cite{shi2021gradient} and \cite{rame2022fishr}, where the classification loss (i.e., local training) and gradient matching (on the central server) are conducted in different places, as such, no gradient-of-gradient information can be obtained. As such, a novel optimization scheme (i.e., aggregation method) should be developed to tackle the problem of domain generalization with non-shared data from multiple domains.

Our formulation is also similar to projecting conflicting gradients (PCGrad) \cite{yu2020gradient}, which is designed for multi-task learning, at a high level. In \cite{yu2020gradient}, the cosine similarity of gradients between two tasks are evaluated. If the value of gradient similarity is negative, PCGrad proceeds to replace one gradient by projecting it onto the normal plane of another gradient.  However, there are two limitations which prevents PCGrad from being applied to our setting, 1) it involves division process when computing cosine similarity of gradient, the training process may not be stable if the gradient vanishes (i.e., values of gradient close to zero) at any local servers, which leads to $0$ divided by $0$; 2) even if there is not gradient vanishing, it is still difficult for PCGrad to be applied in homomorphic encryption based federated learning setting due to the division operation involved \cite{brakerski2014leveled}. Nevertheless, we also show in the experimental section that our proposed method can achieve better performance compared with PCGrad. 

We are also aware that more recently, \cite{yue2022neural} proposed a FL method where clients transmit Jacobian matrices to improve model performance in the non-IID FL setting. While some desired performance was reported in \cite{yue2022neural}, our proposed method is different compared with \cite{yue2022neural} on two folds: 1) we focus on cross-domain FL scenario, 2) we only require the client to tranmit gradient information to the central server, which can be easier to adapt to different FL architectures (based on the standard of federated learning \cite{zhang2022introduction}).

%Besides our proposed method, we also consider another strategy to modify the optimization process of \cite{} to meet the requirement of FL setting as another baseline. The details are introduced in the experiment setting. 

\subsection{Convergence Analysis}

In this section, we conduct convergence analysis of our proposed gradient aggregation method. For simplicity, we assume that two local servers are involved in the federated learning training process, where the gradient sent from two local servers are denoted as $grad_1$ and $grad_2$, respectively. We assume that we first conduct gradient modification on $grad_1$ followed by $grad_2$, if needed. We further denote $\hat{grad}_1$ and $\hat{grad}_2$ as the modified gradients, respectively. We focus on the convergence analysis on local server 1 as a showcase. 

At each update, we have three cases: 
\begin{enumerate}
    \item $\langle grad_1, grad_2 \rangle>0$
    \item $\langle grad_1, grad_2 \rangle<0$ and $\langle \hat{grad}_1, grad_2 \rangle>0$
    \item $\langle grad_1, grad_2 \rangle<0$ and $\langle \hat{grad}_1, grad_2 \rangle<0$
\end{enumerate}

For case 1), there is no need to conduct gradient modification based on our setting, for case 2), we only modify $grad_1$ by keeping $grad_2$ unchanged, for case 3), we modify both $grad_1$ and $grad_2$. Now we are ready to present our analysis. 

\begin{theorem}
	We assume the loss function $\mathcal{L}$ is convex and differentiable, and the gradient of $\mathcal{L}$ is L-Lipschitz with $L > 0$. Then, the model update rule with our proposed gradient modification method will converge to the optimal value.
\end{theorem}

\begin{proof}
If case 1), we can apply gradient descent (e.g., stochastic gradient descent) which leads to a standard deep neural network optimization. 

If case 2), $grad_1$ will be modified as
\begin{equation}
    \hat{grad}_1 = (1-2\lambda)grad_1+2\lambda grad_2,
\end{equation}
and $grad_2$ will keep unchanged. the model parameters will then be updated as 
\begin{equation}
    w^* = w-\frac{\eta}{2}[(1-2\lambda)grad_1+(1+2\lambda)grad_2].
\end{equation}
As we have assumed that $\mathcal{L}$ is Lipschitz continuous, by further denoting $t=\frac{\eta}{2}$, where $t\leq \frac{1}{L}$ based on the Lipschitz continuous property, we can conduct a quadratic expansion of $\mathcal{L}$ around $\mathcal{L}(w)$ and obtain the following inequality:
\begin{eqnarray}\small
  &&  \mathcal{L}(w^*)  \leq  \mathcal{L}(w)+\triangledown \mathcal{L}(w)^\top (w^*-w) +\frac{1}{2}L\|w^*-w\|^2 \nonumber \\
   && \leq  \mathcal{L}(w) + grad_1 (-t[(1-2\lambda)grad_1 + (1+2\lambda) grad_2]) \nonumber \\
    && +  \frac{1}{2}L\|(-t[(1-2\lambda)grad_1 + (1+2\lambda) grad_2])\|^2 \nonumber \\
    && \leq  \mathcal{L}(w) + (2\lambda^2-\frac{1}{2})t\|grad_1 - grad_2\|^2.
\end{eqnarray}
As we can see, since $t>0$, as long as $2\lambda^2-\frac{1}{2}<0$, we can have $\mathcal{L}(w^*)<\mathcal{L}(w)$ which implies that the objective function value strictly decreases with each iteration unless $grad_1 = grad_2$. 

If case 3), $grad_1$ and $grad_2$ will be modified as

\begin{eqnarray}
\hat{grad}_1 & = &  (1-2\lambda)grad_1+2\lambda grad_2,\\
\hat{grad}_2 & = & 2\lambda \hat{grad}_1+ (1-2\lambda) grad_2,
\end{eqnarray}
respectively. Similar to the case 2), we can perform a quadratic expansion around $\mathcal{L}(w)$ as

$$\mathcal{L}(w^*) \approx \mathcal{L}(w) + \triangledown \mathcal{L}(w)^\top (w^* - w) + \frac{1}{2}L \|w^* - w\|^2,$$
where $\triangledown \mathcal{L}(w) = grad_1$, $L$ is the Lipschitz constant, and $w^* - w = \frac{1}{2}(\hat{grad}_1+\hat{grad}_2)$, $\hat{grad}_1  =   (1-2\lambda)grad_1+2\lambda grad_2, \hat{grad}_2 =  2\lambda \hat{grad}_1+ (1-2\lambda) grad_2$.

Now, we substitute the update rule expression for $w^*$ and simplify the inequality using the constraint $t \leq \frac{1}{L}$, which leads to
\begin{equation}
     \mathcal{L}(w^*) \leq \mathcal{L}(w) + (8\lambda^4-\frac{1}{2})t\|grad_1 - grad_2\|^2.
\end{equation}

As long as $8\lambda^4-\frac{1}{2}<0$, we can also have $\mathcal{L}(w^*)<\mathcal{L}(w)$ with each iteration unless $grad_1 = grad_2$. 

In summary, we show that optimal value can be obtained in all the cases. This completes the proof.
\end{proof}
 Our analysis can be extended to the cases where we have multiple local servers. Particularly, one can treat the gradient from the $i$th local server as $grad_1$ the weighted summed gradients from the remaining servers as $grad_2$.  
 
 \subsection{Implementation}
 
 Our proposed gradient aggregation algorithm for privacy-preserving constrained domain generalization (PPDG)  only relies on the gradient information of each client, which is task-agnostic. Intuitively, each client can send gradient to the central server every iteration, however, such mechanism inevitably increases communication burden between central server and the local server, which is not practical in FL.  We thus follow \cite{mcmahan2017communication} to  reduce the computation cost on
each client (i.e., increase the number of iterations of training on each client before sending the information to the central server). In this case, we consider to approximate the gradient  as $\omega_{T}-\omega_0$ (in a form of gradient descent),  where $\omega_0$ denotes the initial parameters of the client model, and $\omega_{T}$ denote the model parameters after $T$ iterations.  We found such strategy to be quite effective on different benchmark datasets.

% \begin{figure}[t] (illustrated in Fig.\ref{fed})
% \centering

% %\setlength{\abovecaptionskip}{0.cm}
% \includegraphics[scale=0.25]{fed.png}
% \caption{Illustration of our proposed federated learning mechanism. }
% \label{fed}
% \end{figure}
 
 %\subsection{Discussion}

% The aforementioned analysis can also be extended to the cases where multiple local servers are involved for training. Assuming that the vanilla aggregated gradients and our proposed aggregated gradients are denoted by $grad_o$ and $grad_{new}$ respectively, and $\lambda$ is sufficiently small as such $cos(grad_o,grad_{new})\geq \frac{1}{2}$, we can obtain the following inequality
% \begin{eqnarray}
%     \mathcal{L}(w^*) & \leq & \mathcal{L}(w) - t grad_o^\top grad_{new} + \frac{1}{2}Lt^2 \|grad_{new}\|^2 \nonumber \\
%     & \leq &  \mathcal{L}(w) - \frac{1}{2}t \|grad_o\|\|grad_{new}\| + \frac{1}{2}t \|grad_{new}\|^2 \nonumber.
% \end{eqnarray}
%ach WILDS dataset is associated with a specific task which caters to a real-world domain generalization problem. As Federated learning framework is usually applied in application scenarios where the data of each client are distributed and data privacy is highly concerned, sometimes  the number of clients could be large,This is also the reason why we do not choose the regular DG benchmarks like PACS/VLCS, as    that Not only that, we also evaluate } our proposed domain generalization algorithm for FL system based on two different medical imaging classification tasks, including skin lesion classification and gray matter segmentation of spinal cord. 

\section{Experiment}
We first evaluate our model on the WILDS benchmark \cite{koh2021wilds}, which contains a variety of datasets  capturing real-world distribution shifts across a diverse range of modalities. We consider three challenging datasets in WILDS benchmark,  namely Camelyon17, Poverty, and FMow where the data are \textit{all with privacy concerns} (e.g., Camelyon17 based on healthcare application, and Poverty, FMow based on satellite imagery which can be related to homeland security), under the Federated learning settings. We consider the \textit{gradient/weight-aggregation-based} FL baselines, including \textbf{FedAvg} \cite{mcmahan2017communication}, \textbf{FedProx} \cite{li2018federated} and \textbf{COPA} \cite{wu2021collaborative}, as well as the domain generalization methods, including \textbf{AgrSum} and \textbf{AgrRand} \cite{mansilla2021domain}, which aim to explore the consensus gradients across domains and can be extended to the FL setting. Besides, we also adopt \textbf{DeepAll} and \textbf{PCGrad} \cite{yu2020gradient} (where PCGrad is designed for multi-task learning but can be extended to the FL setting) as baselines for comparison.

% For simplicity, we assume that the number of local epoch(s) used on each local server is set to 1. %
% \begin{itemize}
%   \item \textbf{DeepAll}: We consider to aggregate all data for training, which is data centralized learning setting, as a baseline for comparison.
%     \item \textbf{FedAvg}: For Federated Averaging (FedAvg) algorithm \cite{mcmahan2017communication}, we perform stochastic gradient descent (SGD) {at the} local server  with the same learning rate and further update the model through gradient averaging process.  
%     \item \textbf{FedProx}: We follow \cite{li2018federated} to introduce a proximal term {on the objective of each local server, mitigating the impact of local-update variation.} 
%   % \item \textbf{BigAug}: We follow \cite{zhang2020generalizing} to conduct data augmentation which aims to improve the generalization capability of trained model for medical imaging classification task. 
%   \item \textbf{AgrSum}: We consider the domain generalization method which is also based on gradient \cite{mansilla2021domain}, where the consensus gradients across domains are summed to form the new gradient.
%   \item \textbf{AgrRand}: We consider a variant of AgrSum by assigning a random value to conflicted gradients across domains. 

% \end{itemize}

Noted that other state-of-the-art non-federated domain generalization baselines are not applicable in our case as they require aggregating data from multiple servers to create a domain shift scenario.

\subsection{Results on Camelyon17 }
The Camelyon17 dataset contains 450,000 scanned patches of breast cancer metastases in lymph-node sections. The data are collected from 5 hospitals, and each hospital can be treated as a single domain. The objective of Camelyon17 is to predict the presence of tumor tissue in the scanned patch. As shown in \cite{koh2021wilds}, the variations in data collection and processing brought from different hospital deployments can greatly degrade the performance of tumor tissue prediction. We follow the setting in \cite{koh2021wilds} for our evaluation, where the training data contain scanned patches from three different hospitals, and the validation and test set consist of data from the rest hospitals. We utilize the training set to train our proposed method and evaluate and validation and test set, respectively. 

%According to the official WILDS-Camelyon17 dataset settings, the train split contains data from 3 hospitals, the validation and test split take the rest two respectively. 

\textbf{Setting. }
We follow the experimental protocols proposed in \cite{koh2021wilds}, using the DenseNet-121 as the network for model training. As for the model training on local server, we set the learning rate to be 0.001, L2-regularization strength to be 0.01, the batch size to be 32 and adopt SGD optimizer with momentum 0.9. We trained the model for 10 rounds. Each round all clients join the training and the local training epochs is set to 1. We choose the epoch with the highest accuracy in validation split, and report the corresponding test accuracy. {We set hyperparameter $\lambda$ of PPDG to 0.1, and }for FedProx baseline, we tune the parameter of the proxy term $\mu$ in a large range and set it to 0.1 where the best performance can be achieved. 

%fine-tune the weight of its proximal term: $\mu$ , set it to 0.1.

\textbf{Results. }
We first conduct performance comparisons with FL methods. 
As can be observed from Table \ref{tab:wilds-camelyon17}, our proposed method can achieve better performance compared with FedAvg and FedProx in a large margin. Such observation is reasonable due to the domain alignment strategy for gradient aggregation, such that shared information among domains can be better exploited. 
%{We also observe that FedProx achieves slightly better performance compared with FedAvg in general. We conjecture the reason that the proximal term can benefit common knowledge learning as it requires the updated model weights to be as close to the original model weights as possible.} 
We also notice that FedProx generally achieves slightly better performance when compared to FedAvg. One possible reason for this improvement could be that the proximal term in FedProx encourages the updated model weights to stay close to the original model weights, which might contribute to enhanced common knowledge learning.

Subsequently, we discuss the performance comparisons with centralized learning. As we can see, FedAvg and FedProx achieve  poorer performance compared with DeepAll baseline in average, which is reasonable due to the data variation across domains. Such results are also consistent with performance in other FL based tasks (e.g., \cite{li2019privacy}). While COPA could achieve better performance compared with FedAvg and FedProx, its generalization capability is still not desired compared with our proposed method. Nevertheless, our proposed method can achieve a competitive performance compared with DeepAll baseline and with better performance on validation set, which is reasonable since our method can be interpreted as conducting domain alignment by mapping the data to the neural kernel space, such that the shareable information across domains can be learned, further bringing benefits to generalization capability. 

Last but not the least, our proposed method can also achieve better performance compared with the gradient based methods AgrSum, AgrRand and PCGrad, which are designed for domain generalization and multi-task learning task but can also be extended to the FL setting. We observe that significant improvement can be achieved by using our proposed, which further shows the superiority of our proposed PPDG. 

%{The result reported in Table \ref{tab:wilds-camelyon17} shows that PPDG can outperforms all other Federated baselines on both validation accuracy and test accuracy. Noted that our method can also achieve competitive performance compared with the centralized DeepAll baseline with just one percentage point margin. We also observed that all gradient-based methods can achieve significant improvement on the validation domain which may suggest that ??? }

\begin{table}[]
\centering
\caption{Results on Camelyon17 dataset.}
\begin{tabular}{lccc}
\hline
\textbf{Method} & Validation. (\%)   & \textbf{Test. (\%)} & \textbf{Average. (\%)}\\ \hline
DeepAll         & 87.4                & 76.8  & 82.1                 \\ \hline
FedAvg          & 80.4                & 70.2  & 75.3                 \\
FedProx        & 80.1                 & 71.4  & 75.8                \\
AgrSum          & 87.4                & 71.1  & 79.3                 \\
AgrRand         & 88.9                & 68.3  & 78.6                 \\
PCGrad          & 85.9                & 70.0  & 77.5                 \\ 
COPA          & 88.0                  & 71.6  & 79.8                 \\ \hline
PPDG            & \textbf{89.0}       & \textbf{73.0}  & \textbf{81.0}        \\ \hline
\end{tabular}
\label{tab:wilds-camelyon17}
\end{table}

\subsection{Results on Poverty}
{The Poverty dataset assembles satellite imagery and survey data (utilized as ground truth) at 19,669 villages from 23 African countries between 2009 and 2016. Poverty is for regression task which aims to predict the real-valued asset wealth index of an area, given its satellite imagery, which is essential for targeted humanitarian efforts in poor regions, especially for much of the developing world where ground-truth measurement of poverty are lacking because of the high field surveys cost. The whole dataset contains 46 different domains: 23 different countries with 2 regions (urban and rural)  for each country. The train split contains 26 domains, the validation and test splits divide the rest 20 domains equally.} We use the train split for model training and evaluate the performance on validation and test splits by computing Pearson correlation (r) between the predicted and ground-truth asset index, the worst group result evaluates the model's generalization ability over the Urban and rural region shift \cite{koh2021wilds}. 

\textbf{Setting. }
{We follow \cite{koh2021wilds} by using the ResNet-18  as the training network, a batch size of 64, and Adam optimizer with an initial learning rate of 0.001 that decays by 0.96 per epoch. We trained the model for 200 epochs and reported the epoch result with highest validation peason-r value along with the corresponding test pearson-r value. {We set the hyper-parameter $\mu$ of FedProx and $\lambda$ of PPDG to 0.1.} As the number of train domains here is large (with 26 domains), we follow \cite{mcmahan2017communication} by choosing a selection ratio 0.5 to randomly select 13 clients (i.e., domains) to join each training round with one local epoch training. }

%after grid search over \{0.01, 0.05 0.1, 0.5, 1.0\}

\textbf{Results. }
{We report the result in Table. \ref{tab:wilds-poverty}. We see that PPDG obtains the highest validation and test performance under both average and worst sections compared with other federated learning based baselines, especially on test split with an improvement of 0.1 ahead over the FedAvg. The centralized DeepAll baseline shows significant better performance, which is reasonable since it can directly access data from all domains and use them during training stage instead of only performing gradient aggregation.}

\begin{table}[]
\centering
\caption{Results on Poverty dataset.} 
\begin{tabular}{lcclcc}
\hline
\textbf{Method} & \multicolumn{2}{c}{Val Pearson $r$ } &  & \multicolumn{2}{c}{\textbf{Test. Pearson $r$}} \\ \hline
                & Average               & Worst         &  & Average            & Worst              \\ \hline
DeepAll         & 0.81                  & 0.52      &  & 0.75               & 0.39  \\ \hline
FedAvg          & 0.71                  & 0.31          &  & 0.69               & 0.13  \\
FedProx         & 0.71                  & 0.29          &  & 0.68               & 0.08   \\
AgrSum          & 0.56                  & 0.21          &  & 0.59               & 0.20   \\
AgrRand         & 0.58                  & 0.28          &  & 0.59               & 0.15    \\
PCGrad          & 0.70                  & 0.34         &  &  0.74         & 0.10              \\ 
COPA            & 0.73                  & 0.25         &  & \textbf{0.80}         & 0.21     \\ \hline
PPDG            & \textbf{0.74}         & \textbf{0.34}          &  & 0.79      & \textbf{0.23}      \\ \hline
\label{tab:wilds-poverty}
\end{tabular}
\end{table}

\subsection{Results on FMow}
{Similar to the Poverty dataset, the FMow dataset also contains satellite images collected from 5 different regions in 16 consecutive years. The task is to predict the land-usage type (62 categories in total such as shopping mall,residential units etc.) from the satellite image. The objective is to generalize the trained model to satellite imagery taken in the future which may be shifted due the infrastructure development across time. Such predictions can contribute to global-scale monitoring of sustainability and economic challenges, aiding policy and humanitarian efforts in applications such as deforestation tracking. We follow the setting in \cite{koh2021wilds} for training, validation and test domain split.  }

\textbf{Settings. }
 %{We followed the official WILDS implementation, training a DenseNet-121 model pretrained on ImageNet for 60 epochs with Adam optimizer. 
 {We follow \cite{koh2021wilds} to train a DenseNet-121 model for the task. The initial learning rate is to set to $10^{-4}$ that dacays by 0.96 per epoch and the batch size is set to 32. We randomly select 5 domains  {to join each training round} with one local training epoch. We set $
\lambda$ of PPDG to 0.05 and $\mu$ of FedProx to 0.01 (we tune $\mu$ in a large range and report the best performance we can achieve).  For evaluation, we  {report} the average accuracy to evaluate the model’s ability to generalize over years, and the worst-case accuracy to measure the model’s generalization performance across regions under a time shift.}

\textbf{Results. }
{We can find in Table \ref{tab:wilds-fmow} that our method achieves the best performance on both test and validation sets among all baselines in terms of the worst-case accuracy, and ranks the second in terms of the average accuracy on validation set. The results further justify the superiority of our proposed method. }
\begin{table}[]
\centering
\caption{Results on FMOW dataset.} 
\begin{tabular}{lcclcc}
\hline
\textbf{Method} & \multicolumn{2}{c}{Val Accuracy (\%)} &  & \multicolumn{2}{c}{Test. Accuracy (\%)} \\ \hline
                & Average               & Worst         &  & Average            & Worst              \\ \hline
DeepAll         & 60.1                  & 50.2          &  & 53.4               & 32.7               \\ \hline
FedAvg          & 57.8                  & 47.7          &  & 52.1               & {32.9}             \\
FedProx         & 56.5                  & 45.2          &  & 50.8               & 31.9               \\
AgrSum          & 52.9                  & 45.7          &  & 47.3               & 27.4               \\
AgrRand         & 53.1                  & 46.8          &  & 47.2               & 28.7               \\
PCGrad          & \textbf{60.1}         & 49.6          &  & 53.7               & 32.8               \\
COPA            & 60.0                  & 47.6          &  & 51.3               & 29.7               \\ \hline
PPDG            & 59.6                  & \textbf{50.4} &  & \textbf{53.8}      & \textbf{33.9}      \\ \hline
\end{tabular}
\label{tab:wilds-fmow}
\end{table}

%\subsection{Parameter analysis}

\subsection{Results on Other Datasets}

Besides only considering datasets from WILDS benchmark with privacy issue, we further evaluation on two other datasets, RMNIST and TerraInc, from Domainbed benchmark \cite{gulrajani2020search}. Specifically, we follow the Domainbed benchmark by using LeNet for RMNIST, and ResNet-18/50 for TerraInc, where we randomly split each source
domain into training and validation set in a ratio of 9:1 to tune the hyperparameter by setting $\lambda=0.001$ for RMNIST, and $\lambda=0.2$ for TerraInc. As for FedProx, we set $\mu=0.1$ where the best performance could be obtained. As we can see, our proposed method can generally outperform other methods, which shows that our proposed method is model agnostic and can be
generalized to various datasets.

\begin{table}[h]
\begin{center}
\caption{Results on RMNIST dataset.}
\begin{adjustbox}{max width=\columnwidth}
\begin{tabular}{lcccccc|c}
\hline
\textbf{RMNIST}  &\textbf{0} &\textbf{15} &\textbf{30} &\textbf{45} &\textbf{60} &\textbf{75} &\textbf{Avg.} \\ \hline \hline
% DeepAll &94.03 &98.7 &98.41 &98.47 &98.53 &94.63 &97.13 \\ \hline
% FedAvg  &82.88 &95.75 &96.63 &96.81 &96.09 &86.3 &92.41 \\
% FedProx$_{0.1}$ &79.64 &94.66 &95.8 &95.88 &94.59 &82.67 &90.54 \\ 
% PCGrad  &86.11 &97.32 &93.77 &95.03 &97.53 &88.43 &93.03 \\
% AgrSum  &72.88 &94.33 &96.35 &96.44 &93.91 &78.81 &88.79 \\
% AgrRand &72.73 &94.32 &96.35 &96.43 &93.8  &79.0  &88.77 \\
% COPA    &83.01 &96.00 &97.36 &97.01 &96.08 &86.63 &93.02  \\
DeepAll &94.0   &98.7 &98.4 &98.5 &98.5 &94.6 &97.1 \\ \hline
FedAvg  &82.9   &95.6 &96.6 &96.8 &96.2 &86.3 &92.4 \\
FedProx &79.6   &94.7 &95.8 &95.9 &94.6 &82.7 &90.5 \\ 
PCGrad  &86.1   &\textbf{97.3} &93.8 &95.0 &\textbf{97.5} &88.4 &93.0 \\
AgrSum  &72.9   &94.3 &96.4 &96.4 &93.9 &78.8 &88.8 \\
AgrRand &72.7   &94.3 &96.4 &96.4 &93.8 &79.0  &88.8 \\
COPA    &83.0   &96.0 &97.4 &97.0 &96.1 &\textbf{88.6} &93.0  \\
\hline
PPDG   &\textbf{84.0} &96.7 &\textbf{97.7} &\textbf{97.8} &97.1 &87.4 &\textbf{93.4} \\ \hline
\end{tabular}
\end{adjustbox}
\end{center}
\label{tab:rmnist_msdg}
\vspace{-10pt}
\end{table}

\begin{table}[h]
\begin{center}
\caption{Results on TerraInc dataset. 
%The baseline results are reported from the corresponding papers. 
}
\begin{adjustbox}{max width=\columnwidth}
\begin{tabular}{lcccc|c}
\hline
\textbf{TerraInc}  &\textbf{Loc.100} &\textbf{Loc.38} &\textbf{Loc.43} &\textbf{Loc.46} &\textbf{Avg.} \\ \hline \hline

\multicolumn{6}{c}{\textbf{Resnet-18}}  \\ \hline
DeepAll &49.8       &31.3   &47.1   &37.2  &41.4 \\
\hline
FedAvg  &46.5       &38.6   &40.2   &27.3  &38.2 \\
FedProx &43.8       &38.1   &39.5   &29.0  &37.6 \\ 
PCGrad  &46.7       &41.1   &40.4   &27.3  &38.9 \\
AgrSum  &\textbf{51.6}       &40.7   &38.6   &35.3           &41.6 \\
AgrRand &50.0       &39.6   &38.3   &\textbf{35.4}  &40.8 \\
COPA    &46.8       &40.6   &42.4   &29.5     & 39.8  \\
\hline
PPDG    &49.0       & \textbf{42.0}  &\textbf{47.6}   &32.7  &\textbf{42.8} \\ \hline

\multicolumn{6}{c}{\textbf{Resnet-50}}  \\ \hline
DeepAll &56.0       &48.0   &54.6   &43.3 &50.5 \\
\hline
FedAvg  &\textbf{59.2}       &48.1   &43.6   &32.9 &45.9 \\
FedProx &50.5       &41.4   &41.0   &32.2  &41.3 \\ 
PCGrad  &57.2       &46.0   &41.7   &33.2  &44.5 \\
AgrSum  &56.7       &47.1   &40.1   &36.3  &45.1 \\
AgrRand &57.4       &46.2   &39.1   &36.9  &44.9 \\
COPA    &59.0       &48.2   &44.6   &33.1  &46.2  \\
\hline
PPDG    &57.2       &\textbf{48.7}  &\textbf{49.9}   &\textbf{37.7}  &\textbf{48.4} \\ \hline
\end{tabular}
\end{adjustbox}
\end{center}
\label{tab:terrainc-msdg-1}
%\vspace{-10pt}
\end{table}

\subsection{Hyperparameter analysis}
\begin{figure}[H]
\centering
\includegraphics[width=0.75\columnwidth]{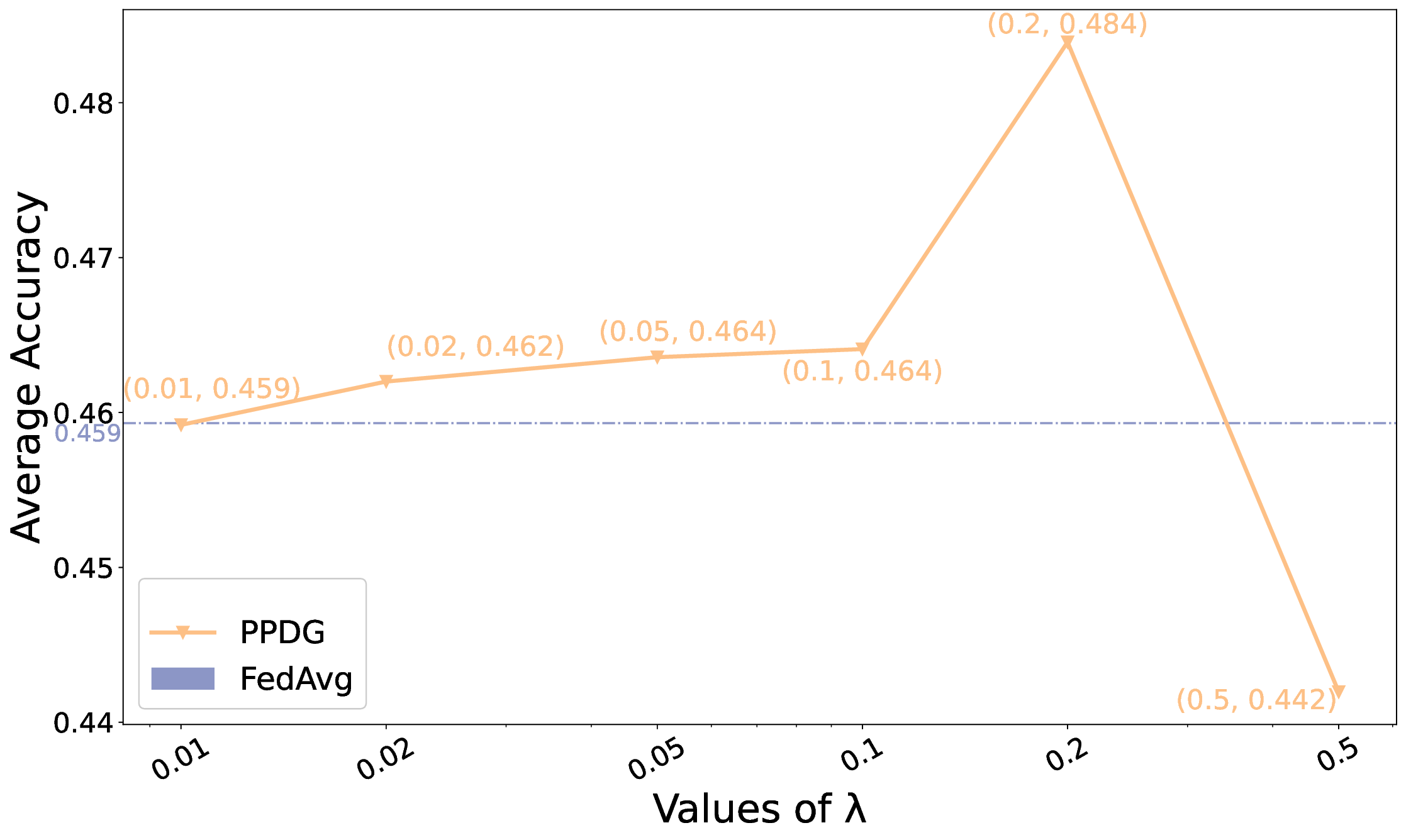}
\caption{Hyperparameter Analysis on $\lambda$ on TerraInc dataset. The dash line denotes the result of FedAvg baseline.}
\label{fig:vis2}
\end{figure}

We now examine the sensitivity of hyperparameter based on the TerraInc dataset. Specifically, our investigation involves assessing the impact of the hyperparameter $\lambda$ over a broad range (i.e., $[0.01,0.02,0.05,0.1,0.2,0.5]$) and the findings of the resulting hyperparameter analysis are presented in the Figure \ref{fig:vis2} (where the results are reported in the format ($\lambda$, average ACC)). The outcomes reveal that the performance is unsatisfactory when $\lambda$ is relatively small, which is reasonable as it may not lead to gradient modification using our proposed approach. Conversely, the performance improves as $\lambda$ increases. Nevertheless, $\lambda$ cannot be excessively large (i.e., $\lambda=0.5$ in our case) due to the risk of non-convergence during optimization, as demonstrated in our theoretical analysis.

\subsection{Ablation study on gradient conflict.} 

We further propose to conduct an ablation study by excluding the consideration of gradient conflict (i.e., always applying gradient alignment loss between two domains) on TerraInc dataset. The findings in Table \ref{tab:grad-conflict} indicate that although the results still surpass the FedAvg baseline, the performance decreases in comparison to the outcome obtained from solely considering gradients that lead to negative transfer. This outcome is reasonable because if the gradients between two domains are excessively similar, an overfitting problem may occur, thereby further increasing the gradient distance between the two domains where negative transfer may transpire.

\begin{table}[H]
\begin{center}
\caption{ Comparison between our PPDG and gradient alignment without considering gradient conflict under the setting on TerraInc dataset where the hyperparameter $\lambda=0.2$.} 

% The results are reported in a p/h format, where p indicates the p-value of the t-test and h = 1 indicates that there is a significant difference between our proposed method and the baseline, while h = 0 indicates no significant difference.}

\begin{adjustbox}{max width=\columnwidth}\label{tab:grad-conflict}
\begin{tabular}{lc}

\hline
\textbf{Method}         &Acc   \\ \hline\hline
FedAvg                  &45.9 \\ \hline
PPDG w/o gradient conflict   &47.0     \\ 
PPDG                    &48.4     \\ \hline
\end{tabular}
\end{adjustbox}
\end{center}
\end{table}

\subsection{Statistical Analysis}
Besides only reporting results based on only average accuracy/Pearson Correlation, we further perform Test of Significance by using paired-sample t-test on Camelyon17 and RMNIST by using p-value at the 5\% significance level. The results are reported in Table \ref{tab:table-pvalue}. From the results, it is evident that h = 1 for all baseline methods, suggesting that it is appropriate to reject the null hypothesis that no difference exists between our proposed approach and the baseline methods. Consequently, we can assert that our proposed method yields significant improvements.

\makeatletter\def\@captype{table}\makeatother
\begin{table}[h]
\begin{center}
\caption{ The results of the t-test, which compares our proposed method with the baseline methods, are presented in a format denoted as p/h. The p-value obtained from the t-test is represented by p. A value of h = 1 indicates a statistically significant difference between our proposed method and the baseline, while a value of h = 0 indicates no significant difference.} 

% The results are reported in a p/h format, where p indicates the p-value of the t-test and h = 1 indicates that there is a significant difference between our proposed method and the baseline, while h = 0 indicates no significant difference.}

\begin{adjustbox}{max width=\columnwidth}\label{tab:table-pvalue}
\begin{tabular}{lccccc}

\hline
\textbf{Method}     &FedAvg       &AgrRand      &ArgSum     &PCGrad     &COPA  \\ \hline\hline
Camelyon17          &0.0024/1     &0.0164/1     &0.0043/1   &1.63e-4/1  &0.0367/1   \\
RMNIST              &0.0033/1     &1.03e-4/1    &1.01e-4/1  &0.043/1    &0.0212/1    \\ \hline
\end{tabular}
\end{adjustbox}
\end{center}
\end{table}

\section{Conclusion}
In this paper, we focus on the domain generalization problem  under the constraint of privacy-preserving settings. In particular, motivated by the theory of kernel embedding on the neural kernel tangent space, we propose a novel gradient aggregation method, which can better extract the shareable information among the data from multiple local servers. We perform experimental studies on various challenging datasets coming from WIDLS and Domainbed benchmarks for classification and regression.  The results justify the effectiveness of our proposed method.  


%\vfill


\bibliographystyle{IEEEtran}
\bibliography{ref}

\end{document}